\documentclass[graybox]{svmult}

\usepackage{mathptmx}       
\usepackage{helvet}         
\usepackage{courier}        
\usepackage{type1cm}        
\usepackage{makeidx}         
\usepackage{graphicx}        
\usepackage{multicol}        
\usepackage[bottom]{footmisc}
\usepackage{hyperref}
\hypersetup{bookmarksopen,bookmarksnumbered,
pdfpagemode=UseOutlines,
colorlinks=true,
linkcolor=blue,
anchorcolor=blue,
citecolor=blue,
filecolor=blue,
menucolor=blue,
urlcolor=blue
}
\usepackage{marvosym}

\usepackage[utf8]{inputenc}
\usepackage[T1]{fontenc}    
\usepackage{upgreek}
\usepackage{amsfonts,amssymb}
\usepackage{bm}
\usepackage{bbm}

\usepackage{amsthm} 
\usepackage{thmtools}
\usepackage{mathtools}
\usepackage{epstopdf}
\usepackage{xspace}
\usepackage[numbers, sort&compress]{natbib}

\usepackage[font=footnotesize]{caption} 
\usepackage[font=footnotesize]{subcaption}
\usepackage{units}
\usepackage{booktabs} 
\usepackage{tabulary}
\usepackage[usenames,dvipsnames]{xcolor} 
\usepackage{diagbox}
\usepackage[ruled,vlined,linesnumbered]{algorithm2e}  
\usepackage{parskip}
\SetKwComment{Comment}{$\triangleright$\ }{}
\usepackage{ifthen,version}
\usepackage{soul}
\usepackage{fixltx2e}
\usepackage{csquotes}
\usepackage{microtype}      
\usepackage{lipsum}
\usepackage{csquotes}
\usepackage{tikz}

\usepackage{enumitem}

\newcommand{\xxnote}[3]{}
\ifx\hidenotes\undefined
  \usepackage{color}
  \renewcommand{\xxnote}[3]{\color{#2}{#1: #3}}
\fi

\newtheoremstyle{hypstyle}
{3pt} 
{3pt} 
{\itshape} 
{} 
{\bfseries} 
{.} 
{.5em} 
{} 

\theoremstyle{hypstyle}

\DeclareMathOperator*{\argmin}{arg\,min}

\newcommand{\argminprob}[1]{\underset{#1}{\argmin}}
\newcommand{\norm}[2]{\left|\left| #1 \right|\right|_{#2}}

\newcommand{\bbm}{\begin{bmatrix}}
\newcommand{\ebm}{\end{bmatrix}}

\newcommand{\Path}[1]{\xi_{#1}}
\newcommand{\cost}[0]{J}
\newcommand{\costFn}[1]{\cost \left( #1 \right)}

\newcommand{\costShot}[0]{\cost_\mathrm{shot}}
\newcommand{\costFnShot}[1]{\costShot \left( #1 \right)}

\newcommand{\costSmooth}[0]{\cost_\mathrm{smooth}}
\newcommand{\costFnSmooth}[1]{\costSmooth \left( #1 \right)}

\newcommand{\costOcc}[0]{\cost_\mathrm{occ}}
\newcommand{\costFnOcc}[1]{\costOcc \left( #1 \right)}

\newcommand{\costObs}[0]{\cost_\mathrm{obs}}
\newcommand{\costFnObs}[1]{\costObs \left( #1 \right)}

\graphicspath{{figs/}}

\makeindex             

\begin{document}
\title*{Autonomous drone cinematographer:\\Using artistic principles to create smooth, safe, occlusion-free trajectories for aerial filming
}
\titlerunning{Autonomous drone cinematographer}
\author{Rogerio Bonatti, Yanfu Zhang, Sanjiban Choudhury, Wenshan Wang, and Sebastian Scherer}
\authorrunning{R. Bonatti, Y. Zhang, S. Choudhury, W. Wang, and S. Scherer}
\institute{R. Bonatti (\Letter) \and S. Choudhury  \and W. Wang \and S. Scherer \at Robotics Institute, Carnegie Mellon University, Pittsburgh, PA, 
\\\email{ rbonatti@cs.cmu.edu}
\\\newline S. Choudhury \\\email{sanjibac@cs.cmu.edu}
\\\newline W. Wang \\\email{wenshanw@andrew.cmu.edu}
\\\newline S. Scherer \\\email{basti @cs.cmu.edu}
\\\newline Y. Zhang \at Advanced Technology Research Division, Yamaha Motor Co., Ltd., Shizuoka, Japan \\\email{zhangya@yamaha-motor.co.jp}
}
%
%
\maketitle







\abstract*{Autonomous aerial cinematography has the potential to enable automatic capture of aesthetically pleasing videos without requiring human intervention, empowering individuals with the capability of high-end film studios. Current approaches either only handle off-line trajectory generation, or offer strategies that reason over short time horizons and simplistic representations for obstacles, which result in jerky movement and low real-life applicability. In this work we develop a method for aerial filming that is able to trade off shot smoothness, occlusion, and cinematography guidelines in a principled manner, even under noisy actor predictions. We present a novel algorithm for real-time covariant gradient descent that we use to efficiently find the desired trajectories by optimizing a set of cost functions. Experimental results show that our approach creates attractive shots, avoiding obstacles and occlusion 65 times over 1.25 hours of flight time, re-planning at 5Hz with a 10s time horizon. We robustly film human actors, cars and bicycles performing different motion among obstacles, using various shot types.}

\abstract{Autonomous aerial cinematography has the potential to enable automatic capture of aesthetically pleasing videos without requiring human intervention, empowering individuals with the capability of high-end film studios. Current approaches either only handle off-line trajectory generation, or offer strategies that reason over short time horizons and simplistic representations for obstacles, which result in jerky movement and low real-life applicability. In this work we develop a method for aerial filming that is able to trade off shot smoothness, occlusion, and cinematography guidelines in a principled manner, even under noisy actor predictions. We present a novel algorithm for real-time covariant gradient descent that we use to efficiently find the desired trajectories by optimizing a set of cost functions. Experimental results show that our approach creates attractive shots, avoiding obstacles and occlusion 65 times over 1.25 hours of flight time, re-planning at 5Hz with a 10s time horizon. We robustly film human actors, cars and bicycles performing different motion among obstacles, using various shot types.}


\section{Introduction}
\label{sec:intro}


\begin{figure}[b]
    \center
    \includegraphics[width=1.0\textwidth]{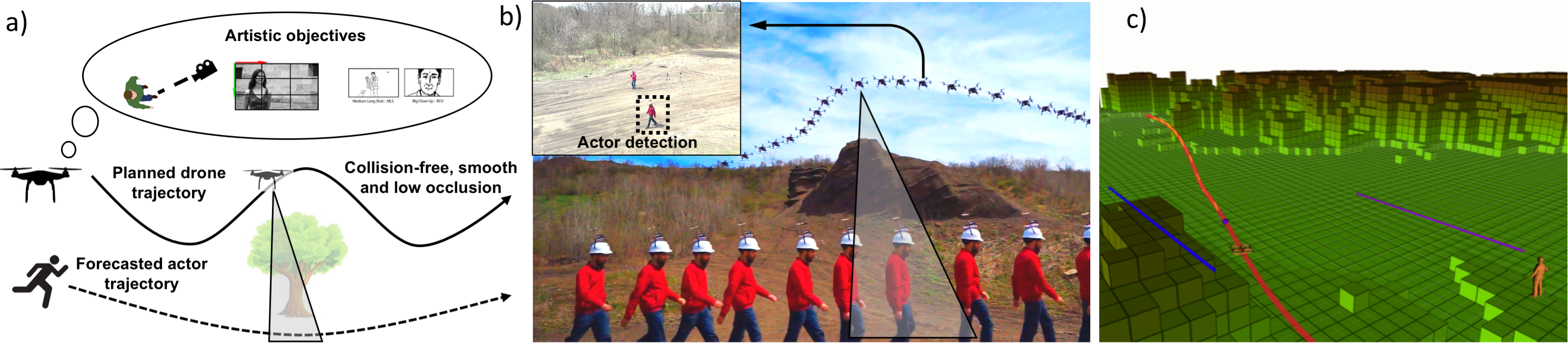}
    \caption{Aerial cinematographer: a) The drone reasons about artistic guidelines, forecasts the actor's motion and plans a smooth, collision-free trajectory while avoiding occlusion. b) Field results produce visually appealing images. The camera detects and tracks the actor, keeping him in the desired screen position. c) Motion planner view, with actor's motion forecast (purple), desired cinematography guidelines (blue), and optimized trajectory (red).}
\end{figure}

Aerial vehicles are revolutionizing the way both professional and amateur film makers capture shots of actors and landscapes, increasing the flexibility of narrative elements and allowing the composition of aerial viewpoints which are not feasible using traditional devices such as hand-held cameras and dollies. 
However, the use of drones for filming today is still extremely difficult due to several motion planning and human-computer interaction challenges. 


Aerial cinematography incorporates objectives from different robotics areas. Similarly to high-speed flight \citep{mohta2018fast,barry2018high}, the drone requires smooth and safe trajectories. Following literature in inspection and exploration, the vehicle reasons about viewpoints. In addition, temporal reasoning plays a major role when following a target, analogous to formation flight \citep{turpin2012trajectory}. Cinematography also needs to consider artistic intent \citep{arijon1976grammar}, using guidelines such as the rule of thirds, scale and relative angles \citep{bowen2013grammar}.

Previous approaches in aerial filming do not address the complete problem in a sufficiently generalizable manner to be used in real-life scenarios. Off-line trajectory generation \citep{roberts2016generating,joubert2015interactive,gebhardt2016airways} cannot be used for most practical situations, and the on-line trajectory generation methods that have been proposed have limitations such as ignoring artistic objectives or only dealing with limited obstacle representations \citep{nageli2017real} (ignoring obstacles altogether in many cases).

Our key insight in this work is that this problem can be efficiently solved in real-time as a smooth trajectory optimization. 
Our contributions in this paper are threefold: (1) we formalize the aerial filming problem following cinematography guidelines for arbitrary types of shots and arbitrary obstacle shapes, (2) we present an efficient optimization method that exploits covariant gradients of the objective function for fast convergence, and (3) for over 1.25 hours of flight time while re-planning, we experimentally show robustness in real-world conditions with different types of shots and shot transitions, actor motions, and obstacle shapes. The supplementary video shows examples of trajectories: \small{\url{https://youtu.be/QX73nBBwd28}}\normalsize.

\section{Problem Formulation}
\label{sec:problem_formulation}



The act of filming encompasses a broad spectrum of behaviors and definitions. For example, a single or multiple cameras can be filming a scene, which can be focused on a landscape, and/or contain one or multiple actors in the frame. In this work we focus on single-camera, single-actor scenarios, which are omnipresent in real-life application, and difficult to execute. Therefore, we have a quadrotor with trajectory $\Path{q}$, coupled with a camera, that films one actor with trajectory $\Path{a}$, where $\Path{}(t)$ maps time $t\in [0,t_f]$ to a configuration. 
At the end of the paper we also discuss how our definitions and algorithms could be extended to other filming scenarios.

\begin{figure}[b]
    \centering
    \includegraphics[width=0.8\textwidth]{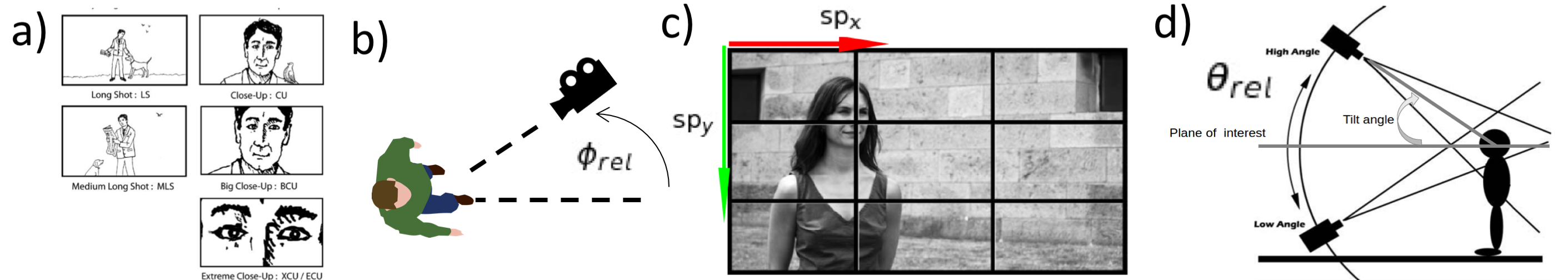}
    \caption{Shot parameters, adapted from \citet{bowen2013grammar}: a) shot scale $ss$ corresponds to the size of the projection of the actor on the screen; b) line of action angle $\phi_{rel} \in [0,2\pi]$; c) screen position of the actor projection $sp_x,sp_y \in [0,1]$; d) tilt angle $\theta_{rel} \in [-\pi,\pi]$}
    \label{fig:cine_params}
\end{figure}

Following literature in cinematography \citep{arijon1976grammar,bowen2013grammar}, we identified a small set of camera positioning parameters that can define a large span of shot types (Figure~\ref{fig:cine_params}). We define \textit{static shots} as shots whose parameters remain static over time, independently of the motion of the actor in the environment, and \textit{dynamic shots} as having time-dependant parameters. 

While filming, a human finds appropriate camera movements based on a set of implicit and explicit costs. We define a generic framework for aerial filming, where the motion planner's objective is find a quadrotor's path $\Path{q}^*$ that minimizes a surrogate cost function $J$, that results in a smooth ($J_{smooth}$), collision-free ($J_{obs}$), occlusion-free ($J_{occ}$) trajectory that follows our artistic shot guidelines ($J_{shot}$) as closely as possible (Eq~\ref{eq:rel_cost}). The optimal trajectory minimizes the total cost $J$ within a finite time horizon $t_f$, while belonging to the set of trajectories $\Xi$ that obey the problem's boundary constraints (Eq~\ref{eq:main_cost}).

\vspace{-5mm}

\small
\begin{align}
\costFn{\Path{q}} &=  \costFnSmooth{\Path{q}} + \lambda_1 \costFnObs{\Path{q}} + \lambda_2 \costFnOcc{\Path{q},\Path{a}} + \lambda_3 \costFnShot{\Path{q},\Path{a}} \label{eq:rel_cost}\\
\Path{q}^* &= \argminprob{\Path{q}\ \in\ \Xi} \quad \costFn{\Path{q}}, \quad \forall t\ \in\ [0,t_f]
\label{eq:main_cost}
\end{align}
\normalsize

\vspace{-10mm}
\section{Related Work} 
\label{sec:related_work}

Camera control in virtual cinematography is typically on through-the-lens control
~\cite{christie2008camera,gleicher1992through,drucker1994intelligent,lino2011director}
but disregards real-world limitations such as robot physics constraints and noisy motion predictions. 
When dealing with arbitrary real-life environments, voxel occupancy maps
and truncated signed distance field (TSDF)
~\citep{newcombe2011kinectfusion}
are common representations, and can supply distance and gradient of a point to nearest object surface. 
Aerial trajectory generation methods~\citep{shim2003decentralized,mellinger2011minimum} are typically designed for aggressive flight and rely on easy to evaluate objective or constraint functions. 
In general domains, we find techniques with more relaxed requirements
~\citep{ratliff2009chomp,schulman2013finding}
. We build on CHOMP~\citep{ratliff2009learning} due to its simple update rule that is amenable to new cost functions.
On aerial filming, we first expose works related to navigation using keyframes. \citet{roberts2016generating} generate off-line trajectories given potentially infeasible human-defined key-frames and \citet{joubert2015interactive} provide a tool for interactive off-line design of camera trajectories. Similarly, \citet{gebhardt2018optimizing} improve a user's time-parametrized trajectory, optimizing it for smooth motions. From user studies, \citet{gebhardt2018optimizing} show that smoothness is key to producing visually-appealing videos. \citet{gebhardt2016airways} also show a smooth trajectory optimization method for user-defined keyframes, including an intuitive user interface. \citet{lino2015intuitive} analytically interpolate between viewpoints while maintaining shot quality. \citet{galvane2017automated} used this method to control quadrotors, but only in obstacle-free environments. Similarly, \citet{joubert2016towards} transition between shots for static actors while not colliding with them, but offer no solution to obstacle and occlusion avoidance. \citet{lan2017xpose} also position the drone using keyframes, which defined by a user on the image itself, not on a global coordinate frame.

On a different line of work, \citet{xie2018creating} generates a set of candidate local camera movements to film landmarks using drones using visual composition rules, and combine the local trajectories into a global smooth camera sequence. 
Closest to our work, we find \cite{nageli2017real,galvane2017directing}. Both methods use high-accuracy indoors motion-capture systems. 
\citet{galvane2017directing} use a search-based planner to find feasible trajectories; however, their system reacts to dynamic targets in a purely reactive manner, resulting in a non-smooth behavior.
\citet{nageli2017real} apply MPC considering occlusion and safety. However, they plan for short time horizons, use simplistic elliptical representations for all obstacles, and use a high-accuracy motion-capture system. It is not clear if the black-box MPC solver that is used is amenable to other obstacle representations and noise in localization.
In contrast, our method works for long time horizons, has a simple unconstrained update rule, operates on TSDFs, has small runtime onboard and can deal with noise in actor motion predictions.

\section{Approach} 
\label{sec:approach}




Unlike previous works that operate either with high-accuracy indoor motion capture systems or precision RTK GPS outdoors, we use only conventional GPS, resulting in high noise for both drone localization and actor motion prediction. Therefore we decided to decouple the motion of the drone and the camera. The camera is mounted on a 3-axis independent gimbal and can place the actor on the correct screen position by visual detection, despite errors in the drone's position. By decoupling camera movement, the trajectory to be optimized becomes $\Path{q}(t) = [x_q(t)\; y_q(t)\; z_q(t)]^T$, assuming that the drone's orientation $\psi_q$ points towards the actor at all times.

\textbf{Designing differentiable cost functions for cinematography}\\
We want trajectories which are smooth, safe, occlusion-free and that follow our artistic guidelines as closely as possible. Following the derivation seen in Section 3 of \citet{zucker2013chomp}, we define a parametrization-invariant smoothness cost that can be expressed as a quadratic function, and an obstacle avoidance cost based on a penalization $c$ of the TSDF. We use obstacle avoidance for both the environment and for the dynamic actor, who for this purpose we represent as a moving sphere. In addition, we  define two more cost functions specifically for cinematography:



\textit{Shot quality}:

\vspace{-7mm}
\small
\begin{eqnarray}
\costFnShot{\Path{q},\Path{shot}} = \frac{1}{t_{f}} \frac{1}{2} \int_0^{t_{f}} ||\Path{q}(t)-\Path{shot}(t)||^2 dt \approx \frac{1}{2(n-1)} Tr(\Path{q}^T A_{shot} \Path{q} + 2\Path{q}^T b_{shot} + c_{shot}) \\
\nabla \costFnShot{\Path{q}} = \frac{1}{n-1}(A_{shot}\Path{q}+b_{shot})
\end{eqnarray}
\normalsize


Written in a quadratic form, it measures the average squared distance between $\Path{q}$ and an ideal trajectory $\Path{shot}$ that only considers positioning via cinematography parameters. $\Path{shot}$ can be computed analytically: for each point $\Path{a}(t)$ in the actor motion prediction, the drone position lies on a sphere centered at the actor with radius calculated via the shot scale, and angles given by $\phi_{rel}$ and $\theta_{rel}$, as in Figure~\ref{fig:cine_params}.

\textit{Occlusion avoidance}:

\vspace{-3mm}
\small
\begin{equation}
\costFnOcc{\Path{q},\Path{a}} = \int_{t=0}^{t_{f}} \int_{\tau=0}^{1}\; c(p(\tau)) \norm{\frac{d}{d\tau} p(\tau)}{} d\tau \, \norm{\frac{d}{dt}\Path{q}(t)}{} dt,
\end{equation}
\normalsize

Even though the concept of occlusion is binary, \textit{i.e}, we either have or don't have visibility of the actor, a major contribution of our work is to define a differentiable cost that expresses a viewpoint's occlusion intensity for arbitrary obstacle shapes.
Mathematically, we define occlusion as the integral of the TSDF cost $c$ over a 2D manifold connecting both trajectories $\Path{q}$ and $\Path{a}$. The manifold is built by connecting each drone-actor position pair in time using the path $p(\tau)$. We then derive the functional gradient:

\vspace{-5mm}
\small
\begin{align}
\nabla &\costFnOcc{\Path{q},\Path{a}}(t) = \nonumber\\ 
& \int_{\tau=0}^{1} \nabla c(p(\tau)) |L| |\dot{q}| \left [I - (\hat{\dot{q}}+\tau(\frac{\dot{a}}{|\dot{q}|}-\hat{\dot{q}}) ) \hat{\dot{q}}^T \right ]
- c(p(\tau)) |\dot{q}| \left [ \hat{L}^T + \frac{\hat{L}^T \dot{L} \hat{\dot{q}}{}^T}{|\dot{q}|} + |L|\kappa^T \right]
d\tau,
\end{align}
\normalsize

where:
\vspace{-7mm}
\small
\begin{center}
$q=\Path{q}(t), \quad a=\Path{a}(t),\quad p(\tau) = (1-\tau)q + \tau a, \quad \hat{v} = \frac{v}{|v|},\linebreak 
\kappa = \frac{1}{|\dot{q}|^2}(I-\hat{\dot{q}} \hat{\dot{q}}^T) \ddot{q}, \quad L = a-q$.
\end{center}
\normalsize
Intuitively, the term $\nabla c(p(\tau))$ is related to variations of the gradient in space, and the term $\tau$ acts as a lever for the gradient. The term $c(p(\tau))$ is linked to changes in path length between camera and actor.

\textbf{Covariant gradient and steepest descent method}\\
Our objective is to minimize the functional $\costFn{\Path{q}}$ (Eq.~\ref{eq:main_cost}). Following a first-order Taylor expansion around the current iteration $i$ using gradient $\nabla J(\Path{q}{}_{i})$, we build Alg.~\ref{alg:optimize}, adapted from \citep{ratliff2009learning}. We follow a direction of steepest descent on the functional cost, using metric $M$, which compounds the quadratic terms coming form both $\costSmooth$ and $\costShot$. $M$ can be seen as an approximation of the Hessian, and only needs to be inverted once, outside of the algorithm's main loop, making the optimization computationally efficient. We follow conventional stopping criteria for descent algorithms, and limit the maximum number of iterations. We use the solution to the previous planning problem concatenated with a straight line segment for future time as the initialization for the next optmization.

\small
\begin{algorithm}[H]
\label{alg:optimize}
\SetAlgoLined
 $M_{inv} \gets (A_{smooth} + \lambda_3 A_{shot})^{-1}$\;
 \For{$i=0,1,...,i_{max}$}{
  \If{$(\nabla J(\Path{q}{}_{i})^T M_{inv} \nabla J(\Path{q}{}_{i}))^2/2 < \epsilon_0$ or $(J(\Path{q}{}_{i}) - J(\Path{q}{}_{i-1})) < \epsilon_1$}{
   \textbf{return} $\Path{q}{}_{i}$\;
   }
  $\Path{q}{}_{i+1} = \Path{q}{}_{i} - \frac{1}{\eta} M_{inv} \nabla J(\Path{q}{}_{i})$\;
 }
 \caption{\small Optimize ($\Path{q}$)}
 \textbf{return} $\Path{q}{}_{i}$\;
\end{algorithm}
\normalsize

\textbf{Gimbal control using the actor's image detection}\\
Our system does not use the actor's state estimation coming from a noisy GPS to control the gimbal to keep the actor within the desired screen position. Instead, we only rely on image feedback. Using the monocular images from the camera, the detection module outputs a bounding box to initialize the tracking process at a higher frame rate. We then use a PD controller to keep the center of the bounding box in the desired screen position.

We selected state of the art algorithms with a good speed-accuracy tradeoff for the vision pipeline. We tested three algorithms: single shot detector (SSD) \cite{liu2016ssd}, Faster R-CNN \cite{ren2015faster}, and YOLO2 \cite{redmon2017yolo9000}. Faster R-CNN performs the best in terms of precision-recall metrics while operating at a reasonable frame rate. We use MobileNet~\cite{howard2017mobilenets} for feature extraction due to its low memory usage and fast inference speed. The per-frame inference is around 300 ms, running on the a Nvidia TX2 computer. We train all models using a mixture of COCO \cite{lin2014microsoft} dataset and our own custom dataset, with a 1:10 ratio, and about 70,000 images in total. We limited the detection categories only to \textit{person}, \textit{car}, \textit{bicycle}, and \textit{motorcycle}, which commonly appear as actors in aerial filming. Finally for actor tracking, we use KCF~\cite{henriques2015high} due to its real-time performance.

\section{Experiments} 
\label{sec:experiments}










\begin{figure}[t]
    \sidecaption
    \includegraphics[scale=0.5]{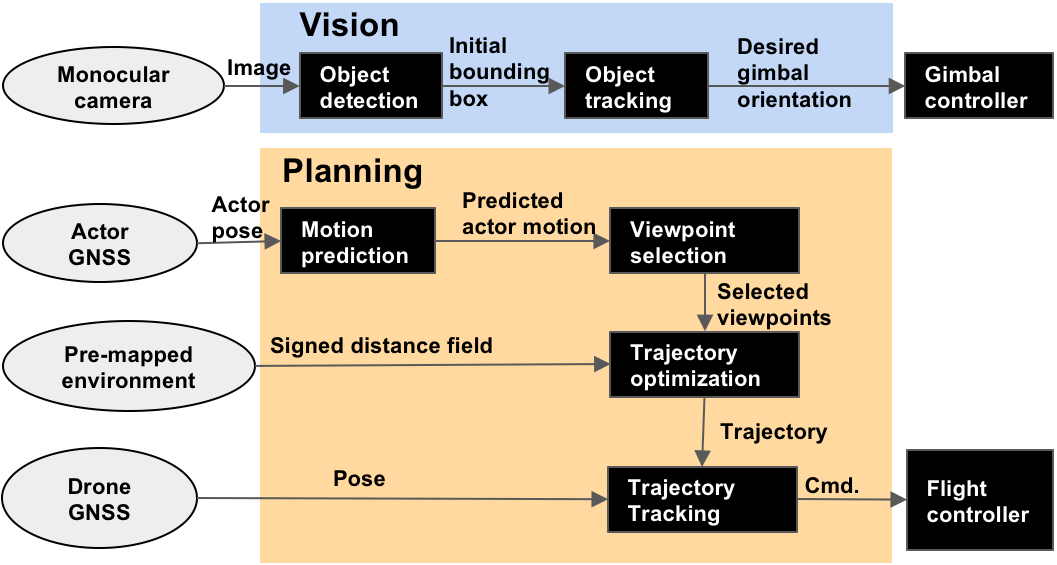}
    \caption{System architecture. The vision subsystem controls the camera orientation using only the monocular image, independently of the planning subsystem. Planning uses the drone's and actor's current location, and the environment to generate trajectories for the flight controller.}
    \label{fig:system2}
\end{figure}

\begin{figure}[b]
    \sidecaption
    \includegraphics[width=0.55\textwidth]{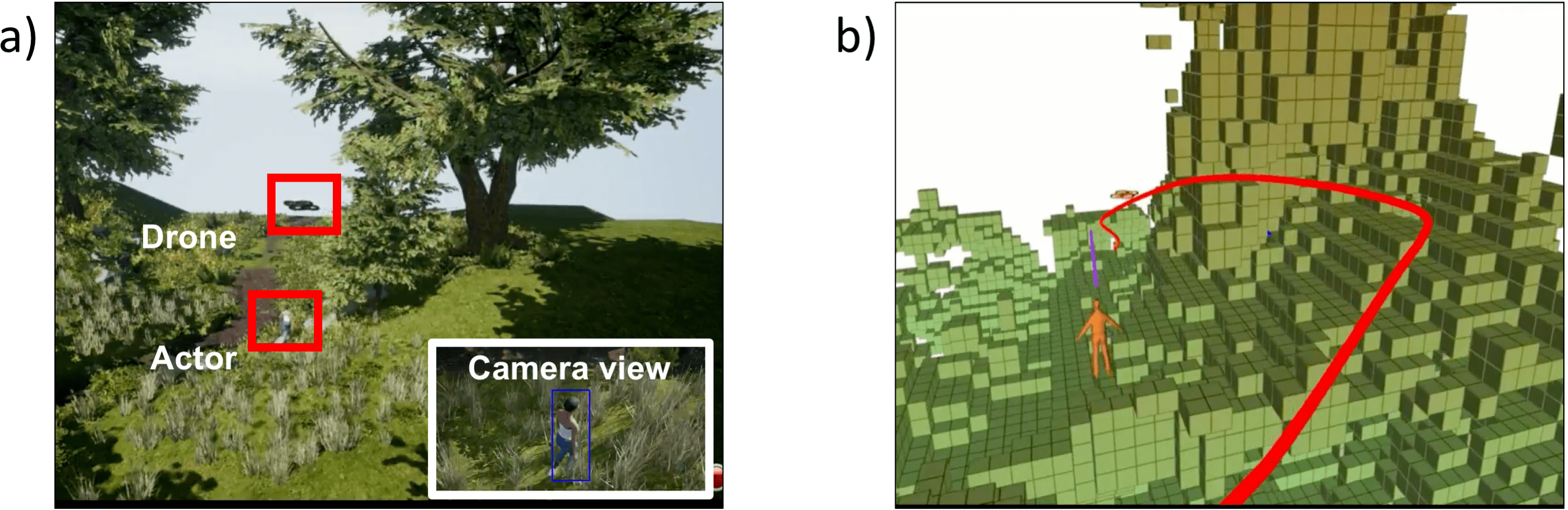}
    \caption{Photo-realistic simulator used to test the system. Third and first-person renderings shown on the left, and occupancy map with drone trajectory shown on the right.}
    \label{fig:simulator}
\end{figure}

\textbf{Systems and experimental setup}\\
Our drone is the DJI M210 model, coupled with a NVIDIA TX2 computer for both vision and planning pipelines (Figure~\ref{fig:system2}).
The actor wears a Pixhawk PX4 module on a hat that sends his pose to the onboard computer via radio communication.
A linear Kalman filter uses these pose estimates to infer his velocity, which is used to predict her trajectory ($\Path{a}$) for the next 10 s. Using a point cloud map of the test site we pre-compute a TSDF map of the region of interest. Re-planning happens at 5 Hz with a 10 s horizon. To simulate the full pipeline and to decide on the relative weights between each cost function (Eq.~\ref{eq:rel_cost}), we built a ROS wrapper to test our software in a photo-realistic environment \cite{airsim2017fsr} (Fig.~\ref{fig:simulator}).

\textbf{Experiments}\\
\textit{(a) Algorithm robustness:}
We evaluated our algorithm performing different types of static and dynamic shots, following different types of actors: humans, cars and bicycles at different speeds and motion types. In total, we collected over 1.25 hours of flight time while re-planning and avoided obstacles and/or occlusion 65 times. The maximum velocity achieved during the tests was of 7.5 m/s. Figure~\ref{fig:res} summarizes the most representative shots, which are also shown in the supplementary video.

\begin{figure}[t]
    \center
    \includegraphics[width=0.9\textwidth]{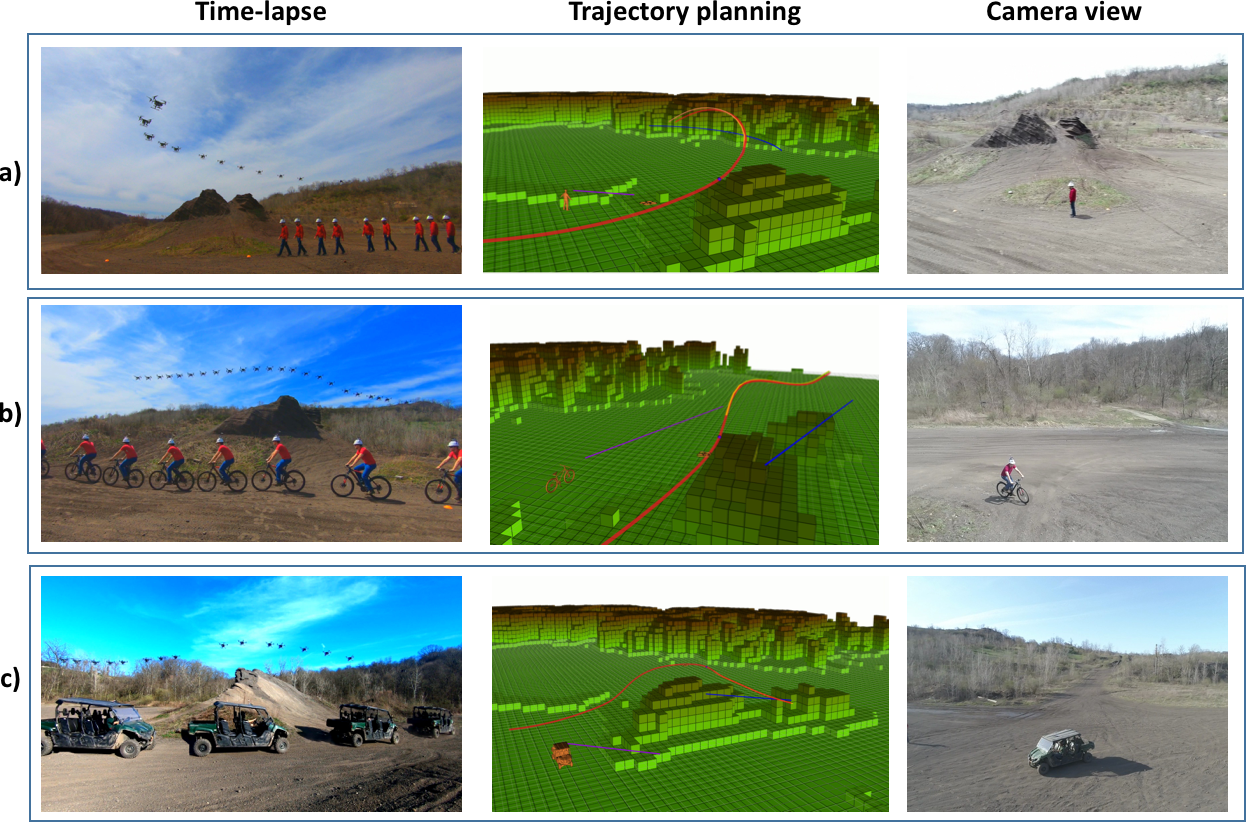}
    \caption{Results: a) Circling shot around person, b) Side shot following biker, c) Side shot following vehicle. The planned trajectory (red) avoids colliding with and being occluded by the mountain, while remaining smooth even under high actor motion prediction noise. The actor's motion forecast is in purple, and the desired artistic shot is in blue.}
    \label{fig:res}
\end{figure}

\begin{figure}[t]
    \sidecaption
    \includegraphics[scale=.24]{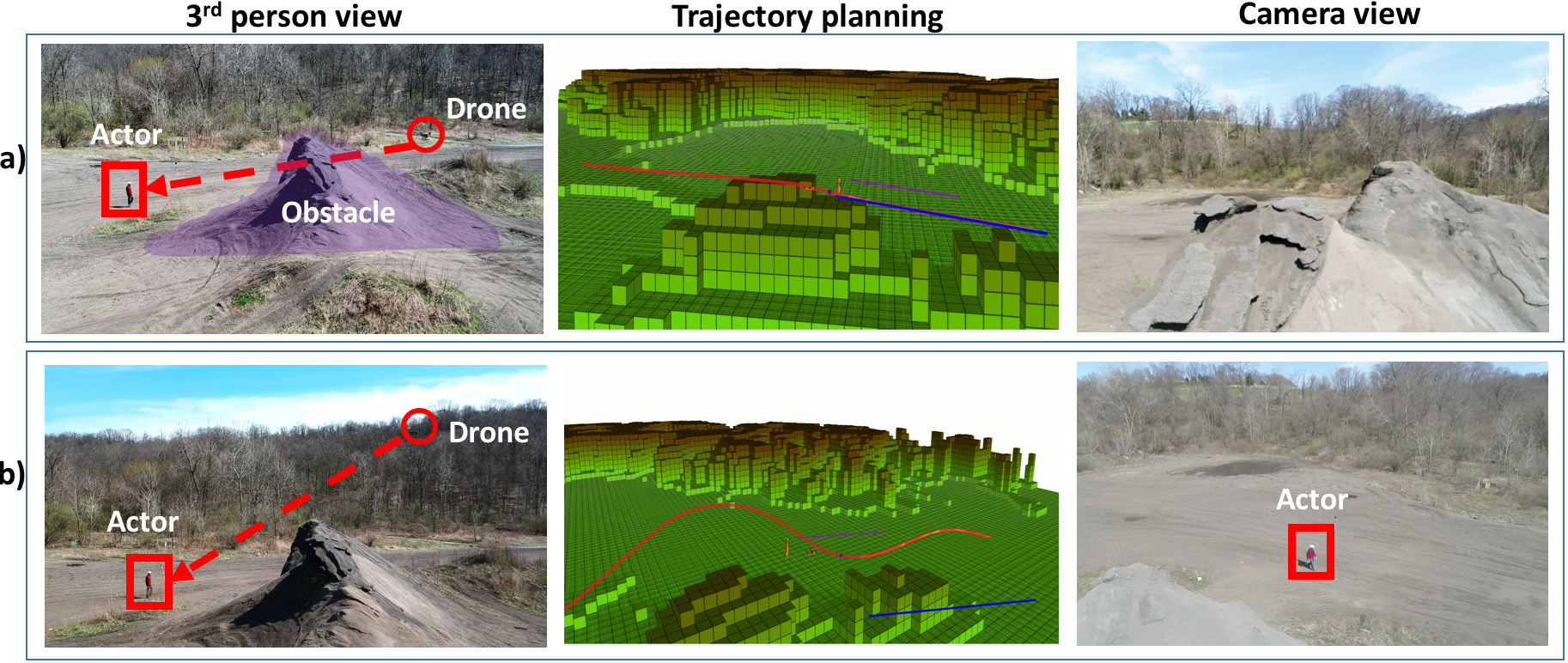}
    \caption{Comparison of planning a) without and b) with occlusion cost function. The occlusion cost function significantly improves the quality of the camera image in comparison with pure obstacle avoidance, for same shot type.}
    \label{fig:occlusion_3}
\end{figure}

\textit{(b) Using occlusion cost function:} For the same shot type, we planned paths with and without the occlusion cost function (Figure~\ref{fig:occlusion_3}). Our proposed occlusion cost significantly improves the aesthetics of the resulting image, keeping the actor on frame while avoiding obstacles.

\begin{figure}[t]
    \sidecaption
    \includegraphics[width=0.64\textwidth]{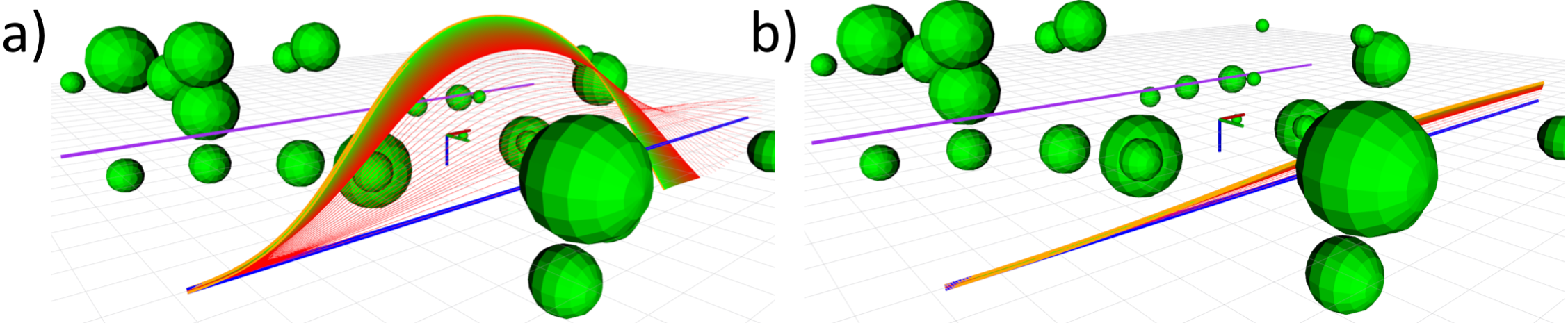}
    \caption{Randomized environment with obstacles to evaluate planner robustness. a) Solution including occlusion cost function, and b) Pure obstacle avoidance. }
    \label{fig:stats}
\end{figure}

\textit{(c) Statistical analysis:}
We evaluate our algorithm on randomized environments (Fig~\ref{fig:stats}), and display results in Table~\ref{tab:1}.


\begin{table}
\caption{Evaluation of motion planner performance in the randomized environment from Fig~\ref{fig:stats}. By adding the occlusion cost function we improve actor visibility over 10\% in comparison with pure obstacle avoidance in environments with 40 spheres. However, by avoiding occlusion the planner also increases the average distance to the desired artistic trajectory. We generated $100$ random configurations for each environment complexity level.}
\label{tab:1}
\begin{tabular}{p{2.5cm}p{2.3cm}p{1.9cm}p{1.9cm}p{1.9cm}}
\hline\noalign{\smallskip}
  & & \multicolumn{3}{c}{\textbf{Num. spheres in environment}} \\
\textbf{Success metric} & \textbf{Cost functions} & $\qquad$ 1 & $\qquad$ 20 & $\qquad$ 40 \\
\noalign{\smallskip}\svhline\noalign{\smallskip}
Actor visibility & $J_{occ}+J_{obs}$ & $99.4 \pm 2.2\%$ & $94.2 \pm 7.3\%$  & $86.9 \pm 9.3\%$  \\
along trajectory & $J_{obs}$ & $98.8 \pm 3.0\%$ & $87.1 \pm 8.5\%$ & $75.3 \pm 11.8\%$ \\
\hline\noalign{\smallskip}
Avg. dist. to $\xi_{shot}$, & $J_{occ}+J_{obs}$ & $0.4 \pm 0.4$ & $6.2 \pm 11.2$ & $10.7 \pm 13.2$ \\
in m & $J_{obs}$ & $0.05 \pm 0.1$ & $0.3 \pm 0.2$ & $0.5 \pm 0.3$  
\end{tabular}
\end{table}

\section{Conclusion and discussion}
\label{sec:conclusion}

In this work we propose and validate a system for autonomous aerial cinematography. Our system is able to execute aesthetically pleasing shots, calculating trajectories that balance motion smoothness, occlusion, and cinematography guidelines. Our experimental results show the algorithm's robustness and speed in a real-life scenarios outdoors, even under noisy actor predictions. In addition, we show that the occlusion cost function we introduced significantly improves the quality of the resulting image, and works for arbitrary obstacle shapes.

There are many key aspects that still need to be solved for us to create a fully autonomous filming system. 
In terms of actor localization, we are currently working on using only visual inputs to identify her position, with no need for GPS. 
In addition, one can incorporate an online-mapping module to the vehicle, which could be integrated with minimal changes to our algorithm, but was out of the scope of this work.
We are also investigating techniques to automatically select the best artistic intent, \textit{i.e}. type of shot, for a particular scenario, taking into account motion cues of the actor and obstacle configurations. 
In the longer term, we also plan to adapt our algorithm to multi-drone and multi-actor configurations by adding a cost function to penalize inter-drone sight. 

\begin{acknowledgement}
We thank Lentin Joseph, Aayush Ahuja, Delong Zhu, and Greg Armstrong for the assistance in field experiments and robot construction. Research presented in this paper was funded by Yamaha Motor Co., Ltd. .
\end{acknowledgement}


\scriptsize{
  \bibliographystyle{plainnat}
  \bibliography{reference}
}
\end{document}